\documentclass[runningheads]{llncs}
\usepackage{graphicx}
\usepackage{hyperref}
\hypersetup{
    colorlinks=true,
    linkcolor=blue,
    filecolor=magenta,      
    urlcolor=cyan
}

\begin{document}
\title{CENSUS-HWR: a large training dataset for offline handwriting recognition}
%
%
\author{Chetan Joshi\inst{1} \and 
Lawry Sorenson\inst{1} \and Ammon Wolfert\inst{1}\and Dr. Mark Clement\inst{1} \and Dr. Joseph Price\inst{1} \and
Dr. Kasey Buckles\inst{2} }
\institute{Brigham Young University, Provo UT 84602, USA,\\
\email{clement@byu.edu}
\and
Department of Economics, University of Notre Dame,
 Notre Dame, IN 46556, USA,\\
\email{kbuckles@nd.edu}
}
%

%
\maketitle              
\begin{abstract}
Progress in Automated Handwriting Recognition has been hampered by the lack of large training datasets.  Nearly all research uses a set of small datasets that often cause models to overfit.  We present CENSUS-HWR, a new dataset consisting of full English handwritten words in 1,812,014 gray scale images. A total of 1,865,134 handwritten texts from a vocabulary of 10,711 words in the English language are present in this collection. This dataset is intended to serve handwriting models as a benchmark for deep learning algorithms. This huge English handwriting recognition dataset has been extracted from the US 1930 and 1940 censuses taken by approximately 70,000 enumerators each year. 
The dataset and the trained model with their weights are freely available to download at \href{http://censustree.org/data.html}{http://censustree.org/data.html}.

\end{abstract}
\section{Introduction}
With the advent of deep learning, researchers have made significant progress in handwriting recognition and transcription. The two most common dataset for the Handwriting Recognition (HWR) are  the IAM \cite{marti2002iam} and RIMES \cite{grosicki12008rimes} dataset. They both contain Latin characters with English and French handwritten sentences respectively. Although these datasets have been useful in creating handwriting models, additional training data is necessary to create more accurate models that can be generalized to more diverse handwritten documents. Limited training data has resulted in very complex models that usually lack explainability and end up overfitting. Such complex models are hard to replicate and need large GPUs to effectively train. This is not ideal for an average researcher or a student as usually they do not have access to such resources.

Additionally, the current datasets for handwriting recognition consist of carefully written text with uniform spacing and do not reflect real-world handwriting. To better train models that are robust to real-world handwriting imperfections, a more diverse and natural dataset is needed. This concern is met by the dataset developed in this research.

\begin{figure}[ht]
    \centering
    \includegraphics[width = 0.6\textwidth]{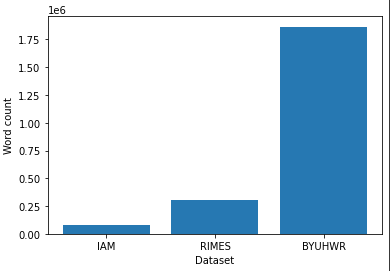}
    \caption{A bar graph of word count comparing the CENSUS-HWR dataset (BYUHWR) with IAM and RIMES dataset. }
    \label{fig:compare_word_count}
\end{figure}

\begin{figure}[ht]
    \centering
    \includegraphics[width = 0.6\textwidth]{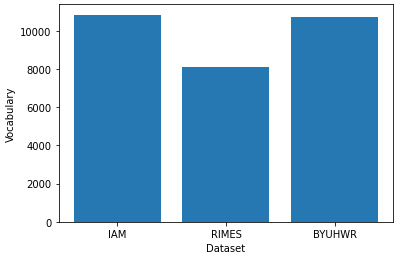}
    \caption{A bar graph of vocabulary size comparing the CENSUS-HWR (BYUHWR) dataset with IAM and RIMES dataset. }
    \label{fig:compare_vocab}
\end{figure}

\begin{figure}[ht]
    \centering
    \includegraphics[width = 0.6\textwidth]{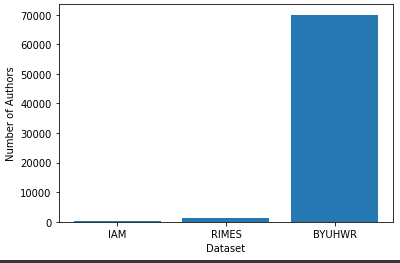}
    \caption{A bar graph of number of authors comparing the CENSUS-HWR (BYUHWR) dataset with IAM and RIMES dataset. }
    \label{fig:compare_num_authors}
\end{figure}

\begin{figure}[ht]
    \centering
    \includegraphics[width = 0.5\textwidth]{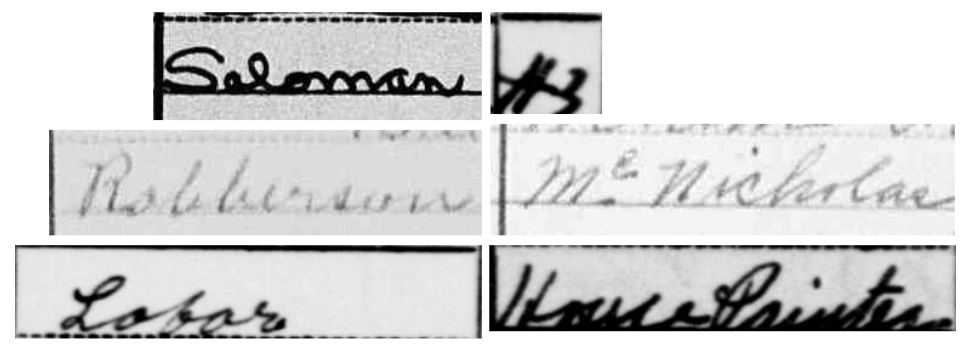}
    \caption{Handwriting data samples from the census images.}
    \label{fig:HWR sample}
\end{figure}

\begin{table}
\caption{Comparison of number of words, vocabulary, number of authors and number of images/forms of CENSUS-HWR (BYUHWR) dataset with the IAM and the RIMES dataset.}\label{tab1}
\centering
\begin{tabular}{|l|l|l|l|}
\hline
 &  {\bfseries IAM} & {\bfseries RIMES} & {\bfseries CENSUS-HWR}\\
\hline
Word count  &  82,227  &  300,000  &  {\bfseries 1,865,134}\\
Vocabulary size  &  {\bfseries 10,841}  &  8,110  &  10,711\\
Image/form count  &  1,066  &  60,000  &  {\bfseries 1,812,014}\\
Authors count  &  400  &  1,300  &  {\bfseries 70,000}\\

\hline
\end{tabular}
\end{table}

\section{Related Work}
The IAM (Institut fur Informatik und Angewandte Mathematik/Department of Computer Science and Applied Mathematics) dataset \cite{marti2002iam} was created to serve as a basis for a variety of offline handwriting recognition tasks. This English Handwriting dataset was particularly useful for recognition tasks where linguistic knowledge beyond the lexicon level is used. Linguistic knowledge can be derived from the underlying corpus \cite{marti2002iam}. To create this dataset, large collections of corpora with different appearances and contents were used. The Lancaster-Oslo/(LOB) \cite{johansson1978manual} collection of 500 English texts, having 2000 words was used as a basis for the dataset. 

The texts in the LOB corpus were quite diverse in nature, ranging from review, religion, biography, and fiction to humour, romance and love stories and adventure. The texts were split into fragments of about 3 to 6 sentences with at least 50 words each which were then printed into forms and several people were asked to write the text printed on the forms using their everyday handwriting. To make the image processing part easy, the writers were asked to use rulers as the guiding lines with 1.5 cm space between them. 

The handwritten words did not contain any compression or deformed words, as they were asked to stop writing if they ran out of space. The handwritten text was written using a ballpoint pen or pencil. These forms were scanned and then labelling was performed. The labels were created by copying the text of the forms and the line feeds were filled manually. For perfect label creation, some manual corrections were made such as deletion, insertion or changes in the text. For text extraction, the skew of the document was corrected, and then a projection method was used to find the position of horizontal lines in the form. With this positional information, the handwritten section was extracted. Later on, the handwritten text was segmented into text lines and each of the text lines was segmented into individual words.

The French RIMES(Reconnaissance et Indexation de données Manuscrites et de fac similÉS / Recognition and Indexing of handwritten documents and faxes) project created a training set for the handwriting recognition and document analysis communities \cite{grosicki12008rimes}. The process of creating this dataset consisted of asking volunteers to write mail in exchange for gift vouchers. The writers were given a fictional identity with their own gender and up to five scenarios one at a time. 

Each of the scenarios consisted of nine realistic themes including change of personal information, information request, opening/closing of the customer account, modification of contracts, complaints, payment difficulties, reminders, damage declaration and destination providers. Each page was scanned and precisely annotated to extract the maximum information which could be useful for evaluation such as layout structure and textual content for transcription. 300,000 handwritten word snippets were extracted from the letters, where each snippet and corresponding ground truth were generated automatically but controlled manually to create an accurate training set. The ground truths obtained were faithful even to the spelling and grammatical errors. 

The other datasets for handwriting recognition are KOHTD \cite{toiganbayeva2022kohtd}, BanglaLekha \cite{biswas_banglalekha-isolated_2017} and the Kurdish dataset \cite{ahmed2021extensive} by Rebin M. Ahmed. The KOHTD, written in the Kazakh language, has 3 different scripts which are Cyrillic, Latin and Arabic providing the diverse script with around 900,000 samples. The BanglaLekha, written in Bengali has around 166,000 samples. Ahmed's Kurdish dataset has 40,095 images written by 390 native writers. However, these datasets also suffer from the same issues, having fewer natural handwritten styles, fewer writers, and fewer training samples.

The datasets described above do not reflect natural handwriting. They are intentionally collected for the aim of handwriting recognition and document analysis. They contain texts that were written carefully in a straight line and special care was taken so the words present in a line or sentence are almost equidistant from each other. Real historical documents are much more messy. For instance, the words in real documents may not be in a straight line, might consist of spelling and grammatical errors, words are overwritten or crossed out and rewritten, the same author might write cursive for a while and then transition to print handwritten words, or the characters might be equidistant in the beginning and then congested at the end due to lack of space. The handwriting community will benefit from a training set that is natural, contains all these errors and is also large enough to avoid over-fitting. The dataset developed in this research meets these criteria and can be used to train models that are more robust to such flaws.

\section{Corpus and Forms}
The CENSUS-HWR dataset has been extracted from the US 1910 census, 1930 census and 1940 census. It includes entries for approximately 92 million people who are enumerated in 1910, 123 million people in 1930 and 132 million people in 1940 as described in National Archives microfilm publications for their respective years: T624, T626, T627. This collection, which is a part of Record Group 29 from the Bureau of the Census, includes the 48 states as well as Alaska, Hawaii, American Samoa, Guam, Consular Services, Panama Canal Zone, Puerto Rico, and the Virgin Islands. The census can be used to identify the place of residence on April 1, 1930, for each person that was enumerated. A manual transcription of these images was created by FamilySearch \cite{FS2002} \cite{FS2012} and Ancestry.com  \cite{An2002} \cite{An2012}. 

The census forms consist of large sheets with rows and columns. The schedules were arranged by states, counties, place and enumeration districts, which were not always filed in sequential order. The census takers were asked to record information about all the people in a household. A county was the basic enumeration unit, which was divided into an enumeration district, one for each enumerator. Once the Census forms were completed, they were sent to the Census Office of the Commerce Department in Washington, D.C. 95-97\% of the population was covered in the schedule. The information on these Federal Censuses was dependent on the informant and the care taken by the enumerator, and hence they are usually reliable. Some of the information in the forms may be incorrect or deliberately falsified.

\begin{figure}[ht]
    \centering
    \includegraphics[width = 0.8\textwidth]{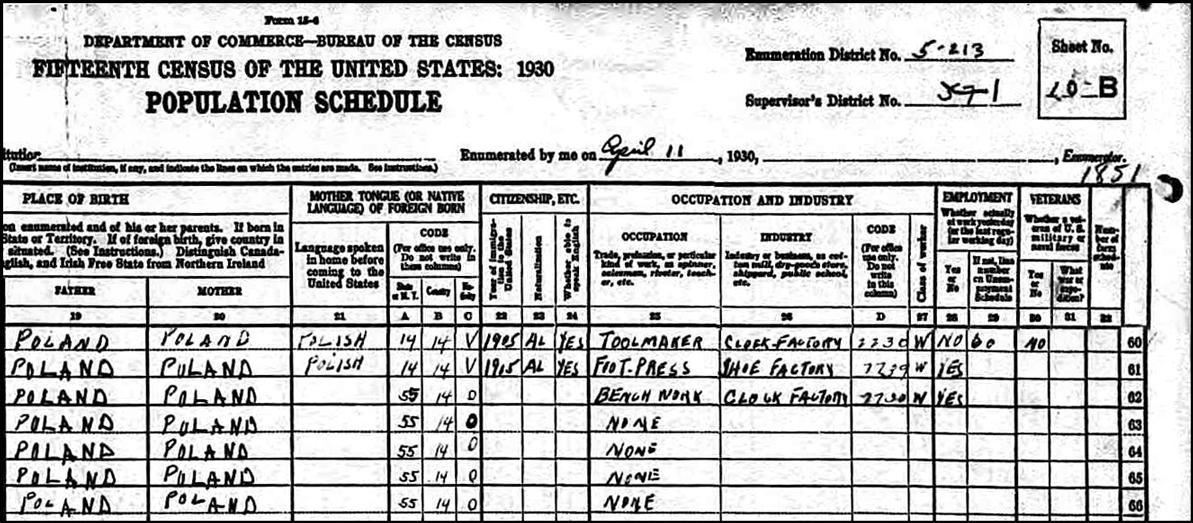}
    \caption{Example of US 1930 Census}
    \label{fig:Census example}
\end{figure}

\section{Text extraction and segmentation}
The handwritten texts have been extracted from the scanned census images using an approach that utilizes scale-invariant feature transform (SIFT) \cite{lowe2004distinctive} and Random sample consensus (RANSAC) \cite{fischler1981random}. SIFT  is generally used to extract the key points of objects from reference images. The objects are recognized in a new image by comparing each feature from the new image to key points from the reference images. It finds matching features based on the Euclidean distance of their feature vectors. Subsets of key points that agree on the object and its location, scale \& orientation in the new image are identified to filter and find better matches. Object matches that pass all the tests are identified as correct with high confidence \cite{lowe2004distinctive}. In our implementation, we use a template (reference) image for each form type; where we label the points of its cells. We compare each sample to its respective template image to find the matching key points.

RANSAC uses repeated random sub-sampling to correlate features in the images with the lines in a template image. The method is tolerant of changes in scale or rotation between the template and image. RANSAC is then used to filter out outliers. \cite{fischler1981random} Based on its findings it becomes easy to align the images and infer the locations of cells in the table. During the segmentation process, each cell of the census is extracted and saved to an image file for that word.

Although the vast majority of census pages were segmented using this method, some pages had severe degradation due to physical damage or image-scanning artefacts and hence this method could not produce segmented word snippets.

\begin{figure}[ht]
    \centering
    \includegraphics[width = 0.8\textwidth]{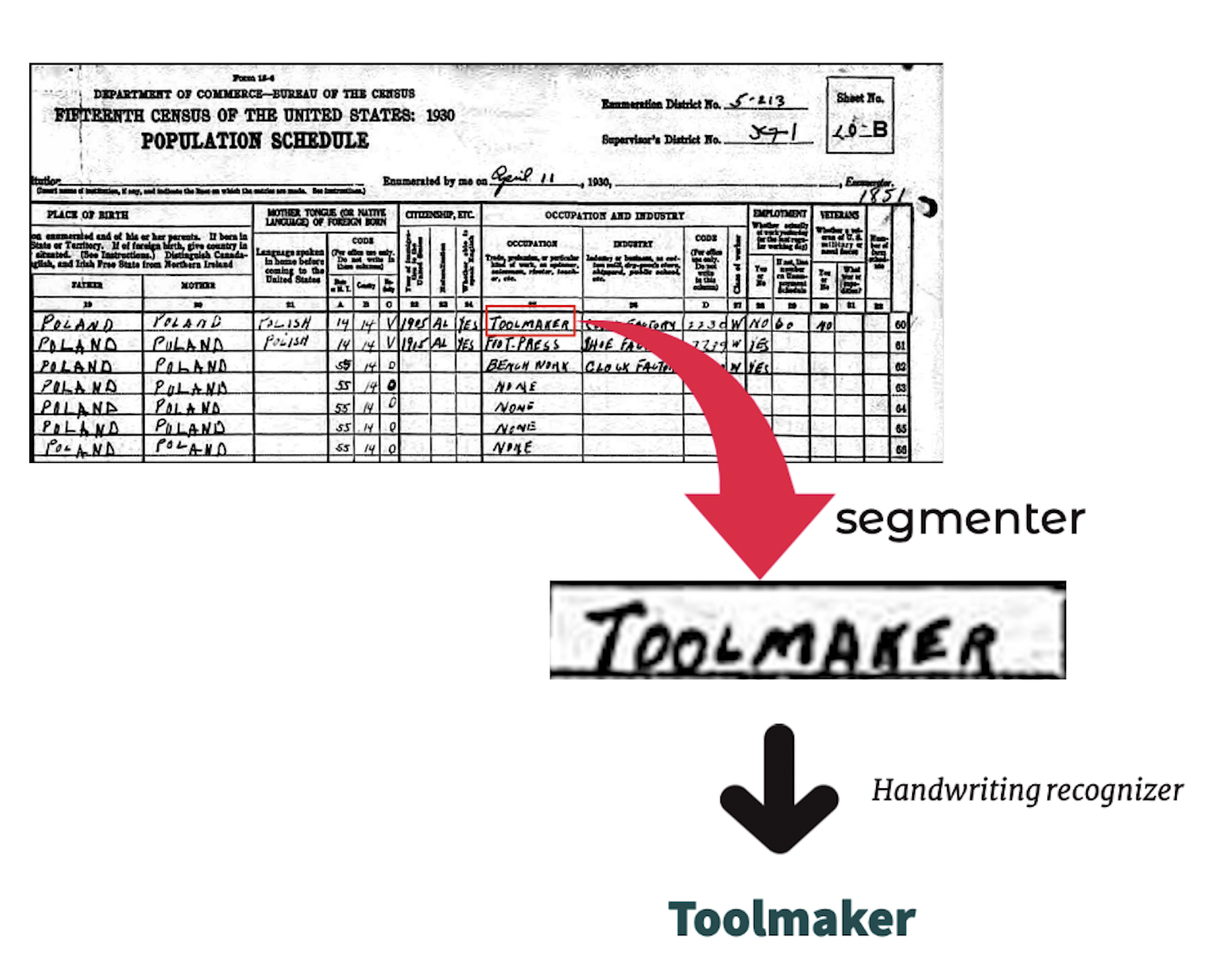}
    \caption{The pipeline of text extraction using segmentation on the census image and then applying handwriting recognition.}
    \label{fig:Pipeline example}
\end{figure}

\section{Labeling}
Each snippet that was extracted from the image by segmentation was assigned a unique image identifier, the row number and the field or column. This information was crucial and helped in matching each snippet with the corresponding FamilySearch transcription. This provided us with a labeled training set of millions of images. This is an unprecedented size for a handwriting training set.

These human transcriptions of the names and dates were then used to train a deep learning model provided along with the CENSUS-HWR dataset.  This automatic transcription model then generated estimated transcriptions of the other fields in the census.  These fields include profession and other fields that were not transcribed by Ancestry and FamilySearch.

\begin{figure}[ht]
    \centering
    \includegraphics[width = 0.8\textwidth]{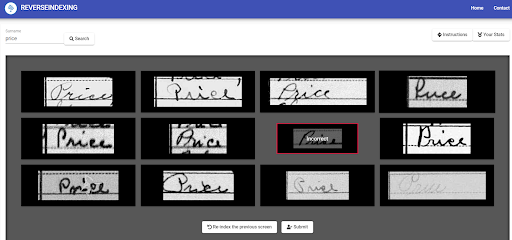}
    \caption{Reverse indexing where humans spot error in transcriptions.}
    \label{fig:reverse_indexing}
\end{figure}

Two tools have been developed to correct transcriptions that were automatically generated by the deep learning model. Both are novel ways to involve humans in transcription and take advantage of the fact that while reading handwriting is a challenging task, recognizing patterns is not. Thus, the first application that was developed is designed to take advantage of the natural human skill to notice cases that are different from those around them. The Reverse Indexing Application groups images based on the transcription produced by the deep learning model. Up to 12 of these images are presented to the user at once with the transcribed version of the name at the top of the screen.  This process allows the user to see multiple versions of the same name and identify any that look different from the rest of the group. All transcriptions are converted to lowercase letters to remove complexity from the process, which allows user to validate only the text of images rather than capitalization.

The images that are marked by the human as likely to be incorrect are loaded into a second application that allows the individuals to submit a free form responses to correct the transcription.

These two applications are used to create labeled training data with an unprecedented level of efficiency and accuracy. This training set will allow us to continue to improve the results of the deep learning model progressively accurate versions of this training set will be shared with other researchers. These applications also have the potential to utilize a large numbers of volunteers as crowd-sourced citizen science projects. 

A version of the Reverse Indexing application is being used on tablets in the prison system for inmates who are willing to provide service hours in exchange for tablet use time. There are over 500,000 tablets in prisons that may be used for Reverse Indexing as this application is rolled out in more states.

\section{Further characteristics of the dataset}
Unlike the intentionally curated English IAM and French RIMES dataset, which were very clean, correctly spaced and without distortions, the CENSUS-HWR dataset contains various distortions, imperfections, mistakes and errors. The images have crossed out words, rewritten or overwritten words, spelling mistakes, congested words, etc. Such a diverse handwritten and natural text truly represents the style people write in a real world setting. This real world representation of handwritten text with various imperfections will allow researchers to develop models which are more robust and can still work efficiently even with imperfect documents. This kind of training data is needed so that researchers can explore new algorithms in the realm of handwriting recognition.

Note that all labels in the dataset are given in lowercase letters, due to the reverse indexing process. This was done to remove complexity from the indexing validation process, since almost all images are lowercase single words with only the first letter capitalized. Since the dataset is validated by crowd-sourced volunteers, it is expected that there are more errors than in professionally annotated datasets. We plan to explore systems for identifying and fixing the mislabeled data in future work, but the current dataset is provided as is.

Table \ref{tab2} details the composition and source of images in the dataset.

\begin{table}
\caption{Sources and types of images in the CENSUS-HWR dataset.}\label{tab2}

\centering
\begin{tabular}{|l|l|l|}
\hline
Source & Type & Count\\
\hline
1930 Census & Surname & 1,178,102 \\
1940 Census & Education Level & 495,183\\
1940 Census & Occupation & 114,301 \\
1940 Census & Industry & 24,428 \\
\hline
\end{tabular}
\end{table}

\section{Handwriting model}
Along with the dataset, we also provide source code and weights for a handwriting model trained on the CENSUS-HWR dataset. The model architecture is based on Bluche et al., 2017 \cite{bluche2017gated}.

The model takes gray scale images as input, which it resizes to 64 x 512 pixels. Original image aspect ratios are preserved, and padding is added as necessary. Six gated convolution blocks reduce the image to 1 x 64 with 512 features. Two bidirectional RNN blocks then map the features to the output character set. The provided model defaults to the same character set as Start, Follow, Read \cite{Wigington_2018_ECCV}. The model is trained with CTC loss. Note that the provided model is limited to predicting 64 characters per image based on the input image size, since most of the dataset images are single words.

We trained preliminary models using 10-fold cross validation on the full dataset without augmentation. The training set was randomly split into training/validation sets with an 85\%-15\% split. Models were trained for twelve epochs, and the weights with the lowest loss on the validation set were saved. Each model was then evaluated on the withheld validation section of the dataset. Table \ref{tab3} contains the results of each test.

\begin{table}
\caption{Character Error Rates (CER) from cross validation tests. The highest and lowest error rates are bolded.}\label{tab3}

\centering
\begin{tabular}{|l|l|}
\hline
Section withheld & CER\\
\hline
1 & 4.6827\% \\
2 & 4.6726\% \\
3 & 4.6830\% \\
4 & \bfseries{4.7115\%} \\
5 & 4.6662\% \\
6 & 4.5915\% \\
7 & 4.6460\% \\
8 & 4.6218\% \\
9 & \bfseries{4.5682\%} \\
10 & 4.6346\% \\

\hline
Mean & \bfseries{4.6478\%} \\
\hline
\end{tabular}
\end{table}

We then trained a model on the full dataset under the same conditions. We estimate the character error rate (CER) of the model as the mean of the cross-validation tests to be 4.6478\%. This model is provided as a proof of concept for using this dataset.

\section{Conclusion}
This paper presents the CENSUS-HWR, which provides a large and natural dataset that represents a diverse variety of natural handwritten styles. This is the largest handwriting dataset with 1.8 million handwritten samples and 70,000 authors that was extracted from the US 1930 census and 1940 census. This dataset is intended to assist the handwriting recognition community in developing robust models. 

While many datasets commonly referenced for handwriting recognition consist of samples with little noise, our dataset reflects the noisiness of most real-world samples. This provides better training for models on a diversity of handwriting styles. And when used in validation and test sets, provides a more rigorous evaluation.

%
%
%
\bibliographystyle{splncs04}
 \bibliography{my_bib}

\end{document}